%% file: arxiv_ver.tex
\documentclass[letterpaper]{article} 
\usepackage[preprint]{aaai2027_arxiv}
\usepackage[hyphens]{url}  
\usepackage{graphicx} 
\usepackage[numbers,sort&compress]{natbib}
\usepackage{caption} 
\usepackage{booktabs}
\usepackage[table]{xcolor}
\usepackage{multirow}
\definecolor{actionlink}{RGB}{91,98,119}
\definecolor{actioncite}{RGB}{105,111,135}
\definecolor{actionurl}{RGB}{78,85,106}
\usepackage[
  colorlinks=true,
  linkcolor=actionlink,
  citecolor=actioncite,
  urlcolor=actionurl
]{hyperref}

\title{\parbox{0.92\textwidth}{\centering
Track4Action: Distilling World-Centric 3D Tracker into Vision-Language-Action Policies\\[0.4em]}}
\author{
  \parbox{0.92\textwidth}{\centering
    Chenyi Wang,\textsuperscript{\rm 1,3*}\quad
    Xinkai Wang,\textsuperscript{\rm 3*}\quad
    Bokai Lin,\textsuperscript{\rm 2,3}\quad
    Jialin Tian,\textsuperscript{\rm 2,3}\quad
    Fucheng Zhang,\textsuperscript{\rm 3}\quad
    Cewu Lu,\textsuperscript{\rm 2,3,4}\quad
    Lixin Yang,\textsuperscript{\rm 2,3,$\dagger$}\\[0.5em]
  }
}
\affiliations{
  \parbox{0.92\textwidth}{\centering
    \textsuperscript{\rm 1}Zhejiang University, China; \textsuperscript{\rm 2}Shanghai Jiao Tong University, China\\
    \textsuperscript{\rm 3}Shanghai Innovation Institute, China; \textsuperscript{\rm 4}Noematrix, China\\
    \texttt{chenyiwang@sii.edu.cn;  siriusyang@sjtu.edu.cn}
  }
}

\newcommand{\method}{Track4Action}

\newcommand{\trackteacher}{Track4World}
\newcommand{\lossalign}{\mathcal{L}_{\mathrm{align}}}
\newcommand{\lossact}{\mathcal{L}_{\mathrm{act}}}
\newcommand{\losstotal}{\mathcal{L}}
\newcommand{\methodname}[1]{\textbf{#1}}
\newcommand{\best}[1]{\textbf{#1}}
\newcommand{\second}[1]{\underline{#1}}

\begin{document}

\maketitle

\begingroup
\renewcommand{\thefootnote}{\fnsymbol{footnote}}
\footnotetext[1]{Equal contribution.}
\footnotetext[2]{Corresponding author.}
\endgroup

\begin{abstract}
Action labels tell a vision-language-action (VLA) policy which robot commands
to imitate, but not how those commands change the 3D world.
The aligned demonstration clip contains this missing supervision because its
$K$ frame transitions record the geometry, motion, visibility, and camera
change produced during the corresponding $K$ actions.
We introduce \method{}, a framework that distills this realized transition from
a frozen world-centric 3D tracker into a current-observation VLA policy.
During training, \trackteacher{} encodes the clip $V_{t:t+K}$ into a pooled
tracker feature.
Learnable track queries infer this feature from current VLA hidden states,
match it in a shared space, and condition a flow-matching action head through a
feature-wise gate.
The tracker feature only defines the alignment target, so neither the clip nor
the tracker is used at deployment.
\method{} reaches 82.3\% on zero-shot LIBERO-Plus, improving the alignment-free
variant by 7.6 points and LaMP by 3.0 points.
It obtains 80.44\% and 81.48\% on the clean and randomized RoboTwin 2.0 splits,
and 67.5\% average success across four physical bimanual tasks, 25.0 points
above the alignment-free variant.
The gains across simulation and physical tasks support action-aligned 3D
tracker features as privileged supervision for tracker-free VLA deployment.
Our project page is available at
\href{https://wing0night.github.io/track4action-project-page/}{wing0night.github.io/track4action-project-page/}.
\end{abstract}

\input{sections/01-introduction}
\input{sections/02-related-work}
\input{sections/03-method}
\input{sections/04-experiments-arxiv}
\input{sections/05-real-world-experiments-arxiv}
\input{sections/06-limitations}
\input{sections/07-conclusion}

\bibliography{aaai2027}


\end{document}

%% file: sections/01-introduction.tex
\section{Introduction}

Vision-language-action (VLA) policies map visual observations and language
instructions directly to robot commands, enabling one policy to learn across
tasks, datasets, and embodiments
\citep{brohan2023rt2,kim2024openvla,black2025pi0}.
This interface provides precise supervision in motor space but leaves the
corresponding world transition implicit.
An action chunk specifies how the robot should move, not which scene points move
with it, whether contact occurs, or how visibility changes during execution.
The same end-effector displacement can traverse free space, transport an object,
or alter the camera view depending on the surrounding 3D structure.
These distinctions determine task success, yet action labels do not encode them.

\begin{figure}[!t]
\centering
\includegraphics[width=1\columnwidth]{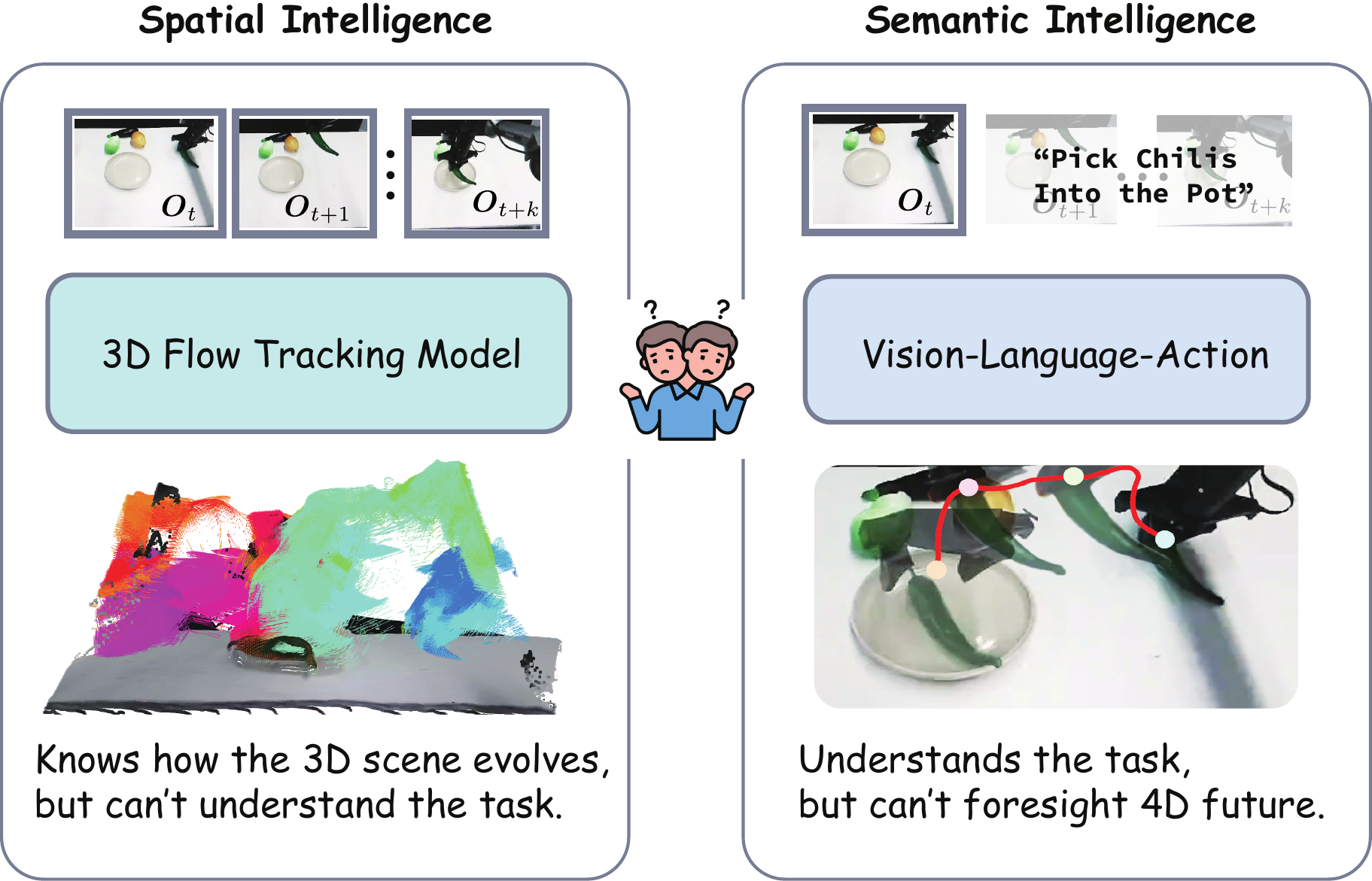}
\caption{\textbf{Track4Action: Unifying 4D Scene Dynamics with Task Understanding.} Flow-tracking models capture how a scene evolves in 4D but lack task awareness, whereas conventional VLA policies understand what to accomplish but do not explicitly model future 4D dynamics. Track4Action bridges these complementary strengths, learning task-conditioned actions with awareness of scene flow.}
\label{fig:teaser}
\end{figure}

Existing work brings future and geometric information into action learning
through complementary routes.
World-model policies predict future images, geometry, or visual features
\citep{zhen20243dvla,pan2025selfcorrecting,su2026worldguidance}, while
geometry-aware policies encode 3D structure or motion priors for action
generation \citep{qu2025spatialvla,wang2026lamp}.
Other methods align policy features with frozen spatial representations or
supervise explicit geometric predictions \citep{li2025spatialforcing,qian2026geopredict}.
These approaches establish that action prediction benefits from structure
beyond current image-language semantics.
Their targets, however, are typically future content, explicit geometric
variables, or spatial features extracted from a single state.
They do not exploit a motion-specialized representation of the exact segment
realized by a demonstrated action chunk.
This gap raises a central question: \emph{Can an action-aligned 3D tracker
supervise a VLA during training, then disappear when the robot acts?}

The demonstration video already contains the required physical evidence.
For an action chunk with $K$ commands, the aligned $K+1$ frames record the scene
before and throughout the corresponding $K$ transitions.
A world-centric 3D tracker organizes these changes into scene geometry, 3D
motion, visibility, and camera cues.
Our key insight is to use the resulting action-aligned tracker feature as
privileged supervision.
The tracker observes the realized transition only while constructing the
training target, whereas the deployed policy must infer the relevant
representation from its current observation and task.

We instantiate this idea in \method{}.
A frozen \trackteacher{} model processes the primary-view clip
$V_{t:t+K}$, where the $K$ frame transitions align with the $K$ actions in the
supervised chunk \citep{lu2026track4world}.
Its scene, motion, and camera tokens are pooled into a world-centric 3D tracker
feature.
Learnable track queries extract a compact representation from the current VLA
hidden states and match the tracker feature in a shared alignment space.
A feature-wise gate then fuses the same queries with the VLA sequence to
condition a flow-matching action head.
The teacher feature only defines the alignment target and never enters the
action head.
Neither the action-aligned clip nor \trackteacher{} is required at deployment.

Two design choices connect privileged supervision to control.
First, temporal alignment ties the tracker target to the same $K$ transitions
supervised by the action chunk, so the target describes an action-specific
world change rather than generic video content.
Second, the track queries act both as the student representation and as an input
to the action head.
The alignment loss therefore shapes an action-facing representation rather than
a detached auxiliary decoder.
This separation supplies world-transition supervision without making the
realized clip or tracker a deployment dependency.

Across LIBERO-Plus, RoboTwin 2.0, and physical bimanual manipulation, the same
distillation interface supports different observations, embodiments, and
action spaces.
On zero-shot LIBERO-Plus, \method{} reaches 82.3\%, improves all seven
perturbation categories over the alignment-free variant, and exceeds LaMP by 3.0
points in the benchmark-provided average.
It achieves 80.44\% and 81.48\% on the clean and randomized splits of the
50-task RoboTwin 2.0 benchmark.
On four physical tasks, \method{} reaches 67.5\% average success, outperforming
the alignment-free variant by 25.0 points while remaining tracker-free at
execution.

Our contributions are summarized as follows:
\begin{itemize}
    \item We formulate world-centric 3D tracker feature distillation for VLA
    policies, turning the realized action segment into privileged supervision
    for a current-observation policy.
    \item We introduce an action-facing track-query pathway whose features both
    match the tracker target and condition the action head through gated fusion,
    while retaining tracker-free deployment.
    \item We evaluate the same interface under zero-shot visual shifts,
    50-task bimanual simulation, and physical manipulation, reaching 82.3\% on
    LIBERO-Plus and 67.5\% average success on four real-world tasks.
\end{itemize}

%% file: sections/02-related-work.tex
\section{Related Work}

\begin{figure*}[!t]
\centering
\includegraphics[width=0.98\textwidth]{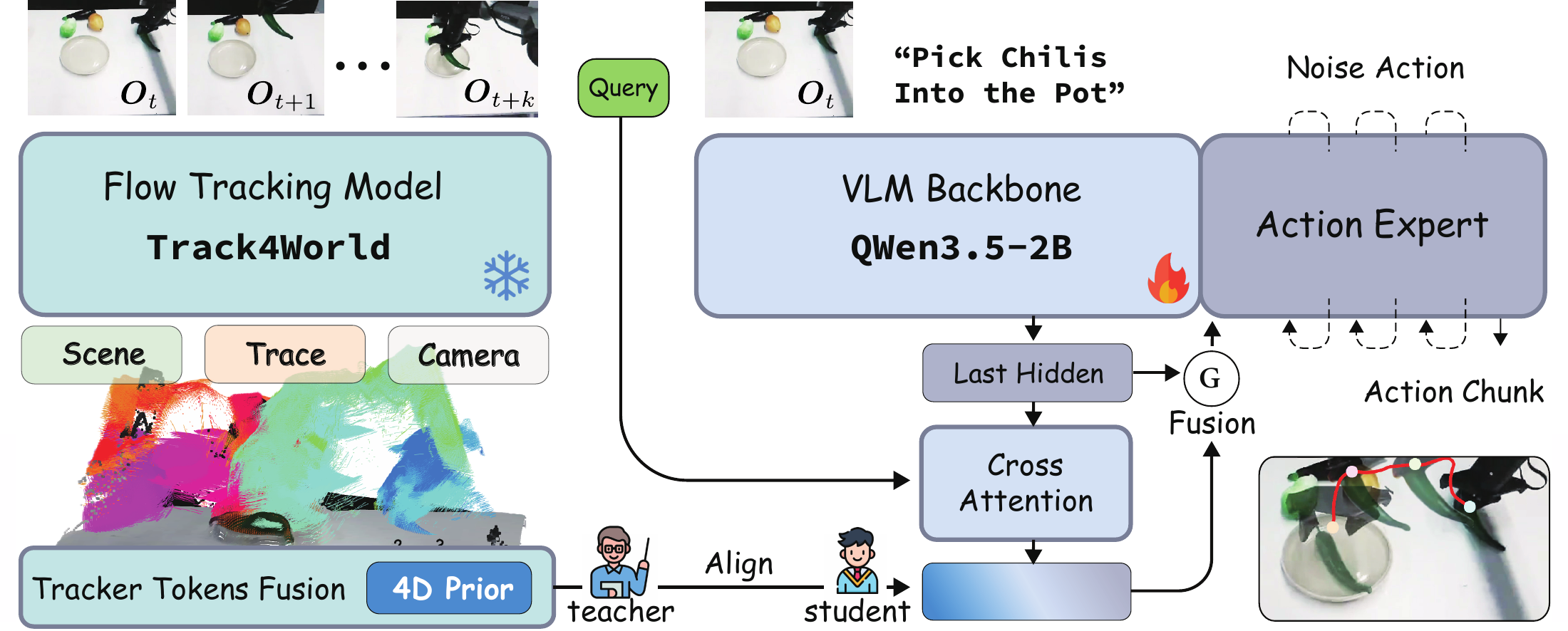}
\vspace{2pt}
\caption{\textbf{World-centric 3D tracker feature distillation in \method{}.}
During training, frozen \trackteacher{} encodes the action-aligned clip
$V_{t:t+K}$ into a pooled target built from scene, motion, and camera tokens.
Track queries extracted from current VLA hidden states match this target and
are fused back into the VLA sequence through a feature-wise gate to condition
the flow-matching action head.
The $K$ actions correspond to the $K$ transitions between the $K+1$ teacher
frames.
Only the current-observation policy path is retained at deployment.}
\label{fig:framework}
\end{figure*}

\paragraph{Vision-language-action policies.}
VLA policies couple pretrained vision-language representations with robot
action decoders and learn generalist control from heterogeneous demonstrations
\citep{brohan2023rt1,brohan2023rt2,oneill2024openx,ghosh2024octo}.
Representative designs include autoregressive action tokenization in OpenVLA
and continuous flow matching in $\pi_0$
\citep{kim2024openvla,black2025pi0}.
Although these policies differ in backbone and action representation, their
primary learning signal is an expert-action likelihood or denoising objective.
This signal specifies an executable command but not the scene geometry and
motion produced by its execution.
\method{} preserves the current-observation VLA interface and instead
supervises the action-facing representation, making the approach compatible
with different VLA backbones and action decoders.

\paragraph{Future and geometric supervision for action learning.}
Prior work enriches action learning through two complementary forms of
supervision.
Future-prediction methods model upcoming images, geometry, visual features, or
motion to guide action generation
\citep{zhen20243dvla,pan2025selfcorrecting,su2026worldguidance,sun2026vlajepa,lin2026hifvla}.
Geometry-aware methods encode 3D structure, learn scene flow, align policy
features with frozen spatial representations, or predict explicit geometric
variables
\citep{qu2025spatialvla,wang2026lamp,li2025spatialforcing,qian2026geopredict}.
\method{} shares their goal of strengthening action representations but uses a
different supervisory object.
A frozen world-centric tracker reads the exact multi-frame segment realized by
the demonstrated action chunk, and the current-observation student predicts its
pooled feature.
The resulting target is neither a deployment input nor an output that the robot
must reconstruct at test time.

\paragraph{3D tracking and motion representations.}
Point tracking provides long-range motion representations with less dependence
on appearance \citep{doersch2023tapir,karaev2024cotracker}.
Robot-learning methods consequently use tracks or flow as an intermediate
between video and control.
Track2Act converts predicted 2D tracks into end-effector plans, TraceGen models
future motion in a cross-embodiment 3D trace space, and 3DFlowAction generates
object flow for action planning
\citep{bharadhwaj2024track2act,lee2025tracegen,zhi2025flowaction}.
JOPAT jointly denoises visual latents, 2D tracks, visibility, and actions
\citep{guan2026jopat}.
These approaches expose trajectories or flow as a prediction, plan, or
action-generation objective.
In contrast, \trackteacher{} provides a pretrained representation of dense
world-centric 3D trajectories, visibility, scene motion, and camera parameters
\citep{lu2026track4world}.
\method{} pools this representation over the realized action segment and
distills it into current-context track queries.
The policy neither reconstructs individual tracks nor executes the tracker at
deployment, preserving the original VLA control interface.

%% file: sections/03-method.tex
\section{Method}
\label{sec:method}

\method{} uses the realized world transition in a demonstration to supervise
the representation that predicts the corresponding action chunk.
Figure~\ref{fig:framework} separates the training-only tracker branch from the
current-observation policy retained at deployment.
We first define the action-aligned training sample, then construct a
world-centric tracker target, distill it into track queries, and connect the
queries to flow-matching action generation.

\subsection{Problem Formulation}
At time $t$, the policy receives a language instruction $l$, current RGB
observations $I_t$, and optional robot state $s_t$.
It predicts the normalized action chunk
\begin{equation}
A_t^\star=(a_t,\ldots,a_{t+K-1})\in\mathbf{R}^{K\times d_a}.
\end{equation}
During training, the same sample also provides a primary-view clip
$V_{t:t+K}=(o_t,\ldots,o_{t+K})$.
The $K$ actions span the $K$ transitions between these $K+1$ frames.
We use this temporal correspondence to attach a world-transition target to the
same segment supervised by $A_t^\star$.
The deployed policy predicts
\begin{equation}
\widehat{A}_t=\pi_\theta(l,I_t,s_t).
\end{equation}
The clip $V_{t:t+K}$ is absent from the policy inputs.
It only constructs privileged supervision during training and is unavailable
at deployment.

\subsection{World-Centric Tracker Supervision}
Action coordinates specify a robot command but not its world-space effect.
The same end-effector displacement can move through free space, establish
contact, or transport an object depending on scene geometry.
We therefore construct the missing target from the realized transition that
accompanies the action chunk.

Let $\mathcal{T}$ denote the frozen \trackteacher{} model.
Given the action-aligned clip, it estimates world-centric 3D trajectories
together with scene, motion, visibility, and camera cues.
We pool these outputs into a compact tracker feature:
\begin{equation}
f_t^{\mathrm{trk}}
=
\mathrm{sg}\!\left[
\mathrm{Pool}\!\left(\mathcal{T}(V_{t:t+K})\right)
\right],
\end{equation}
where $\mathrm{sg}$ stops gradients into \trackteacher{}.
Unlike pixel reconstruction, $f_t^{\mathrm{trk}}$ summarizes the 3D structure
and motion recovered from the realized action segment.
Because the clip covers exactly the transitions induced during
$A_t^\star$, the target describes the world change associated with that action
chunk rather than an unrelated future event.

\subsection{Distilling Tracker Features into the VLA}
The student must infer the tracker feature without observing the realized
transition.
The VLA backbone first encodes the current images and instruction as
\begin{equation}
H_t=\mathcal{E}_\theta(l,I_t).
\end{equation}
We introduce learnable track queries $Q$ that read $H_t$ through
cross-attention:
\begin{equation}
Z_t=\mathcal{D}_\theta(Q,H_t).
\end{equation}
The resulting sequence forms an action-facing bottleneck between the
multimodal backbone and the action head.
We pool $Z_t$ and compare it with $f_t^{\mathrm{trk}}$ in a shared feature
space:
\begin{equation}
\lossalign
=
\left\|\bar{z}_t-\bar{f}_t^{\mathrm{trk}}\right\|_2^2,
\end{equation}
where $\bar{z}_t$ and $\bar{f}_t^{\mathrm{trk}}$ denote the two representations
mapped to the shared alignment space.
The alignment objective trains the queries to infer the likely 3D transition
from the current scene and task context.
The same queries also condition action generation.
Tracker supervision therefore shapes a representation consumed by the policy
instead of an isolated auxiliary decoder.

\subsection{Action Generation and Tracker-Free Deployment}
Tracker alignment can affect control only if the action decoder uses the
aligned representation.
We therefore fuse $Z_t$ with the VLA sequence through a learned feature-wise
gate to form the action condition $H_t^{\mathrm{cond}}$.
The gate preserves the semantic context in $H_t$ while exposing the distilled
tracker cues to the action head.
The teacher feature $f_t^{\mathrm{trk}}$ never enters
$H_t^{\mathrm{cond}}$.

We use a flow-matching action head \citep{lipman2023flowmatching} and optimize
its standard conditional velocity objective $\lossact$.
The full objective is
\begin{equation}
\losstotal
=
\lambda_{\mathrm{act}}\lossact
+\lambda_{\mathrm{align}}\lossalign.
\end{equation}
The action loss teaches the policy what to execute.
The alignment loss teaches its action-facing representation which world
transition accompanies that execution.

At deployment, the policy computes $H_t$, $Z_t$, and
$H_t^{\mathrm{cond}}$ from $(l,I_t,s_t)$ and generates an action chunk from
noise.
We discard the action-aligned clip, frozen tracker, tracker target, and
alignment branch.
The deployed policy retains the track-query and gated-fusion pathway without
running \trackteacher{}.

%% file: sections/04-experiments-arxiv.tex
\section{Experiments}
\label{sec:experiments}
We organize the evaluation around three questions:
\begin{itemize}
    \item \textbf{Comparison:} Does \method{} outperform competitive general,
    video-based, trace-guided, and 3D-aware VLA policies
    (Sections~\ref{sec:libero_results}--\ref{sec:robotwin_results})?
    \item \textbf{Tracker distillation:} Does the complete tracker-distillation
    pathway improve control over its alignment-free counterpart
    (Section~\ref{sec:tracker_effect})?
    \item \textbf{Challenging settings:} Where do its gains persist across
    long-horizon tasks, seven zero-shot shifts, 50-task bimanual simulation,
    and physical manipulation
    (Sections~\ref{sec:libero_results}--\ref{sec:real_world})?
\end{itemize}
We answer these questions with in-distribution LIBERO, zero-shot LIBERO-Plus,
RoboTwin 2.0, and tracker-free execution on a physical bimanual platform.

\subsection{Experimental Setup}
Our simulation evaluation covers LIBERO \citep{liu2023libero}, LIBERO-Plus
\citep{fei2025liberoplus}, and RoboTwin 2.0 \citep{chen2025robotwin}.
LIBERO measures in-distribution performance across Spatial, Object, Goal, and
Long suites.
LIBERO-Plus evaluates the same training regime without adaptation under seven
perturbation dimensions.
For both benchmarks, we train one policy across the four original LIBERO
suites.
We report the benchmark-provided \emph{Avg} for LIBERO-Plus rather than
recomputing an unweighted mean of its columns.
RoboTwin 2.0 tests 50 bimanual tasks under clean and randomized conditions.

\method{} uses a Qwen3.5-2B VLA backbone and a DiT-B flow-matching action head
\citep{peebles2023dit}, for 3.3B parameters in total.
On LIBERO, the policy receives agent-view and wrist-view RGB images plus an
8-D robot state and predicts 7-D delta end-effector actions with $K=8$.
On RoboTwin, it receives one external and two wrist views without robot state
and predicts 14-D bimanual joint and gripper commands with $K=16$.
The alignment-free variant retains the VLA backbone, action head, model scale,
and rollout interface but trains without the tracker-distillation pathway.
We also keep the data, optimizer, and training schedule fixed.
This controlled comparison directly tests our central hypothesis: whether the
complete tracker-distillation pathway improves control under matched data,
optimization, model scale, and rollout settings.
We report online task success rate for every benchmark.

\subsection{LIBERO Results}
\label{sec:libero_results}
\input{sections/table/tab_libero}
\paragraph{Overall comparison.}
Table~\ref{tab:libero_results} shows that \method{} reaches 97.0\% average
success, 0.1 points above $\pi_{0.5}$ and 1.3 points above
$\mathcal{F}_1$.
It obtains the highest scores on Object and Long while remaining within one
point of the best result on Goal.

\paragraph{Where does tracker supervision help?}
Removing tracker alignment lowers the average from 97.0\% to 94.2\%.
The 2.8-point average gap is concentrated on LIBERO-Long, where success drops
from 95.8\% to 86.2\%.
In contrast, Spatial, Object, and Goal change by only 1.0, 0.4, and 0.2 points.
The 9.6-point Long gain suggests that tracker supervision is most useful when
the policy must preserve task-relevant state across multiple stages, rather
than uniformly increasing already saturated short-horizon scores.
Because standard LIBERO leaves little headroom, we next test whether this gain
survives fully zero-shot distribution shifts.

\subsection{Zero-Shot Robustness on LIBERO-Plus}
\label{sec:libero_plus_results}
\input{sections/table/tab_libero_plus}
\paragraph{Overall comparison.}
All policies in Table~\ref{tab:libero_plus_results} train only on original
LIBERO demonstrations and receive no adaptation to LIBERO-Plus.
\method{} reaches 82.3\% average success, 3.0 points above LaMP at 79.3\%.
It ranks first on Camera, Light, Noise, and Layout and second on Robot,
Language, and Background.
The model therefore places within the top two on every perturbation dimension
instead of trading robustness in one category for another.

\paragraph{Consistency across zero-shot shifts.}
Tracker alignment improves all seven dimensions over the alignment-free
variant and raises the official average from 74.7\% to 82.3\%.
The largest gains occur on Camera at 18.3 points, Noise at 9.0 points, and
Robot at 8.7 points.
The absence of a category-level regression shows that the 7.6-point average
gain is not driven by a single favorable perturbation.
LaMP remains stronger on Robot, Language, and Background, which identifies
three settings where additional robustness is still available.

\subsection{Bimanual Evaluation on RoboTwin 2.0}
\label{sec:robotwin_results}
\input{sections/table/tab_robotwin_res}

\paragraph{Overall comparison.}
As shown in Table~\ref{tab:robotwin_results}, \method{} reaches 80.44\% under
clean conditions and 81.48\% under domain randomization.
It exceeds the strongest reported baseline, Motus w/o Pretrain, by 2.88 and
4.48 points.
The margins over X-VLA are 7.64 and 8.64 points.
The RoboTwin results extend the comparison from single-arm Cartesian control
to three-view observations and 14-D bimanual actions.

\paragraph{Behavior under randomization.}
The success rate of \method{} increases by 1.04 points from clean to randomized
conditions.
Over the same change, Motus decreases by 0.56 points and X-VLA changes by only
0.04 points.
Consequently, the margin over Motus grows from 2.88 to 4.48 points under domain
randomization.
The 1.04-point increase shows that the clean-setting advantage is retained when
RoboTwin randomizes the scene.


\subsection{Cross-Benchmark Effect of Tracker Distillation}
\label{sec:tracker_effect}
The alignment-free controls answer the second question through a system-level
ablation of the complete tracker-distillation pathway.
Under matched data, optimization, model scale, and rollout settings, the
pathway raises average success by 2.8 points on LIBERO and 7.6 points on
LIBERO-Plus.
On RoboTwin 2.0, the gains are 39.12 points under clean conditions and 41.58
points under randomization.
The smallest margin appears on saturated in-distribution LIBERO, while the
larger margins appear under zero-shot shifts and a different bimanual action
space.
Together with the 9.6-point gain on LIBERO-Long, the largest improvements occur
in settings that place greater demands on world state and motion.
The consistent gains across observation shifts, task horizons, and action
spaces support the complete tracker-distillation pathway as effective
privileged supervision for control.

%% file: sections/table/tab_libero.tex
\begin{table}[t]
\caption{\textbf{LIBERO success rate (\%).}
We train a single policy across all four suites and report the mean over the
four suite scores.
Bold marks the best result in each column, and underlining marks the second best.}
\label{tab:libero_results}
\centering
{\small
\setlength{\tabcolsep}{5pt}
\resizebox{\columnwidth}{!}{%
\begin{tabular}{lccccc}
\toprule
Method & \multicolumn{5}{c}{LIBERO} \\
\cmidrule(lr){2-6}
 & Spatial & Object & Goal & Long & Avg \\
\midrule
\multicolumn{6}{l}{\textbf{\textit{General VLA}}} \\
OpenVLA \citeyearpar{kim2024openvla} & 84.7 & 88.4 & 79.2 & 53.7 & 76.5 \\
$\pi_0$ \citeyearpar{black2025pi0} & 96.8 & 98.8 & 95.8 & 85.2 & 94.2 \\
$\pi_{0.5}$ \citeyearpar{black2025pi05} & \best{98.8} & 98.2 & \best{98.0} & \underline{92.4} & \underline{96.9} \\
GR00T N1 \citeyearpar{bjorck2025grootn1} & 94.4 & 97.6 & 93.0 & 90.6 & 93.9 \\
\midrule
\multicolumn{6}{l}{\textbf{\textit{Latent-Action VLA}}} \\
UniVLA \citeyearpar{bu2025univla} & 96.5 & 96.8 & 95.6 & 92.0 & 95.2 \\
villa-X \citeyearpar{chen2025villax} & 97.5 & 97.0 & 91.5 & 74.5 & 90.1 \\
\midrule
\multicolumn{6}{l}{\textbf{\textit{Video-Based VLA}}} \\
mimic-video \citeyearpar{pai2025mimicvideo} & 94.2 & 96.8 & 90.6 & -- & 93.9 \\
WorldVLA \citeyearpar{cen2025worldvla} & 87.6 & 96.2 & 83.4 & 60.0 & 81.8 \\
$\mathcal{F}_1$ \citeyearpar{lv2025f1} & \underline{98.2} & 97.8 & 95.4 & 91.3 & 95.7 \\
\midrule
\multicolumn{6}{l}{\textbf{\textit{Flow/Trace-Guided VLA}}} \\
FlowVLA \citeyearpar{zhong2025flowvla} & 93.2 & 95.0 & 91.6 & 72.6 & 88.1 \\
TraceVLA \citeyearpar{zheng2024tracevla} & 84.6 & 85.2 & 75.1 & 54.1 & 74.8 \\
\midrule
\multicolumn{6}{l}{\textbf{\textit{3D-Aware VLA}}} \\
SpatialVLA \citeyearpar{qu2025spatialvla} & 88.2 & 89.9 & 78.6 & 55.5 & 78.1 \\
\methodname{\method{} w/o Align.} & 94.0 & \underline{99.2} & 97.4 & 86.2 & 94.2 \\
\methodname{\method{}} & 95.0 & \best{99.6} & \underline{97.6} & \best{95.8} & \best{97.0} \\
\bottomrule
\end{tabular}
}
}
\end{table}

%% file: sections/table/tab_libero_plus.tex
\begin{table*}[!t]
\caption{\textbf{LIBERO-Plus zero-shot OOD success rate (\%).}
All models train on original LIBERO demonstrations and receive no additional
training for the seven perturbation dimensions.
\emph{Avg} follows the official benchmark aggregation.
Bold marks the best result in each column, and underlining marks the second best.}
\label{tab:libero_plus_results}
\centering
{\small
\setlength{\tabcolsep}{3pt}
\begin{tabular}{lcccccccc}
\toprule
Method & Camera & Robot & Language & Light & Background & Noise & Layout & Avg \\
\midrule
UniVLA \citep{bu2025univla} & 1.8 & 46.2 & 69.6 & 69.0 & 81.0 & 21.2 & 31.9 & 42.9 \\
OpenVLA \citep{kim2024openvla} & 0.8 & 3.5 & 23.0 & 8.1 & 34.8 & 15.2 & 28.5 & 15.6 \\
OpenVLA-OFT \citep{kim2025openvlaoft} & 56.4 & 31.9 & 79.5 & 88.7 & 93.3 & 75.8 & 74.2 & 69.6 \\
$\pi_0$ \citep{black2025pi0} & 13.8 & 6.0 & 58.8 & 85.0 & 81.4 & 79.0 & 68.9 & 53.6 \\
$\pi_0$-FAST \citep{pertsch2025fast} & \underline{65.1} & 21.6 & 61.0 & 73.2 & 73.2 & 74.4 & 68.8 & 61.6 \\
WorldVLA \citep{cen2025worldvla} & 0.1 & 27.9 & 41.6 & 43.7 & 17.1 & 10.9 & 38.0 & 25.0 \\
LaMP \citep{wang2026lamp} & 64.5 & \best{69.6} & \best{88.2} & \underline{95.3} & \best{97.4} & 76.9 & 73.8 & \underline{79.3} \\
\midrule
\methodname{\method{} w/o Align.} & 58.3 & 52.0 & 84.3 & 92.7 & 90.2 & \underline{79.4} & \underline{76.1} & 74.7 \\
\methodname{\method{}} & \best{76.6} & \underline{60.7} & \underline{87.1} & \best{96.3} & \underline{96.1} & \best{88.4} & \best{78.7} & \best{82.3} \\
\bottomrule
\end{tabular}
}
\end{table*}

%% file: sections/table/tab_robotwin_res.tex
\begin{table*}[!htbp]
\caption{\textbf{RoboTwin 2.0 success rate (\%) under clean and randomized
settings.}
The Motus w/o Pretrain results are from \citet{bi2026motus}.
Averages are computed over all 50 tasks.
Bold and underline denote the best and second-best results within each setting.}
\label{tab:robotwin_results}
\centering
{\small
\setlength{\tabcolsep}{2pt}
\begin{tabular}{p{3.0cm}cccccccccc}
\toprule
\multirow{2}{*}{Simulation Task} & \multicolumn{2}{c}{$\pi_{0.5}$~\citeyearpar{black2025pi05}} & \multicolumn{2}{c}{X-VLA~\citeyearpar{zheng2025xvla}} & \multicolumn{2}{c}{Motus w/o Pretrain~\citeyearpar{bi2026motus}} & \multicolumn{2}{c}{\methodname{\method{} w/o Align.}} & \multicolumn{2}{c}{\methodname{\method{}}} \\
\cmidrule(lr){2-3}\cmidrule(lr){4-5}\cmidrule(lr){6-7}\cmidrule(lr){8-9}\cmidrule(lr){10-11}
 & Clean & Rand. & Clean & Rand. & Clean & Rand. & Clean & Rand. & Clean & Rand. \\
\textbf{Model Size} & \multicolumn{2}{c}{3.3B} & \multicolumn{2}{c}{0.9B} & \multicolumn{2}{c}{8B} & \multicolumn{2}{c}{3.3B} & \multicolumn{2}{c}{3.3B} \\
\midrule
\emph{Place Dual Shoes} & 12\% & 7\% & \best{79\%} & \best{88\%} & \second{78\%} & \second{80\%} & 28\% & 28\% & \best{79\%} & 79\% \\
\emph{Move Stapler Pad} & 16\% & 18\% & \second{78\%} & \best{73\%} & 49\% & 37\% & 17\% & 17\% & \best{79\%} & \second{68\%} \\
\emph{Stack Blocks Two} & 48\% & 56\% & 92\% & 87\% & \best{96\%} & \best{94\%} & 74\% & 77\% & \second{94\%} & \second{93\%} \\
\emph{Scan Object} & 42\% & 38\% & 14\% & 36\% & 42\% & 50\% & \second{55\%} & \second{57\%} & \best{88\%} & \best{86\%} \\
\emph{Place Object Stand} & 74\% & 65\% & 86\% & 88\% & \second{91\%} & \second{93\%} & 70\% & 65\% & \best{96\%} & \best{96\%} \\
\emph{Place Fan} & 25\% & 36\% & \second{80\%} & 75\% & 77\% & \second{85\%} & 37\% & 35\% & \best{89\%} & \best{94\%} \\
\emph{Move Pillbottle Pad} & 33\% & 29\% & 73\% & 71\% & \second{83\%} & \second{83\%} & 36\% & 32\% & \best{99\%} & \best{93\%} \\
\emph{Pick Dual Bottles} & 10\% & 6\% & 47\% & 36\% & \second{58\%} & \second{68\%} & 7\% & 6\% & \best{84\%} & \best{92\%} \\
\emph{Blocks Ranking Rgb} & 43\% & 35\% & \second{83\%} & 83\% & \best{92\%} & \second{88\%} & 10\% & 8\% & 79\% & \best{91\%} \\
\emph{\ldots{} (50 tasks)} &  &  &  &  &  &  &  &  &  & \\
\emph{Turn Switch} & 5\% & 6\% & 40\% & \second{61\%} & \best{69\%} & 60\% & 18\% & 13\% & \second{59\%} & \best{72\%} \\
\emph{Pick Diverse Bottles} & 5\% & 3\% & \second{58\%} & 36\% & 53\% & \second{62\%} & 10\% & 13\% & \best{73\%} & \best{81\%} \\
\emph{Place Bread Basket} & 48\% & 56\% & \second{81\%} & 71\% & 73\% & \second{83\%} & 23\% & 21\% & \best{89\%} & \best{89\%} \\
\emph{Stack Blocks Three} & 15\% & 16\% & 6\% & 10\% & \best{71\%} & \best{76\%} & 31\% & 27\% & \second{68\%} & \second{63\%} \\
\emph{Put Bottles Dustbin} & 12\% & 9\% & \best{74\%} & \second{77\%} & 36\% & 33\% & 3\% & 3\% & \second{47\%} & \best{81\%} \\
\emph{Place Can Basket} & 19\% & 25\% & \second{49\%} & 52\% & 46\% & \second{62\%} & 39\% & 31\% & \best{69\%} & \best{66\%} \\
\emph{Stamp Seal} & 36\% & 23\% & \second{76\%} & \second{82\%} & \best{80\%} & \best{88\%} & 12\% & 9\% & 74\% & 73\% \\
\emph{Hanging Mug} & 3\% & 3\% & \second{23\%} & \best{27\%} & 14\% & \second{10\%} & 12\% & 7\% & \best{26\%} & \best{27\%} \\
\emph{Handover Block} & 18\% & 19\% & \second{73\%} & \second{37\%} & 34\% & 15\% & 16\% & 21\% & \best{75\%} & \best{77\%} \\
\emph{Stack Bowls Three} & 33\% & 35\% & 76\% & \best{86\%} & \best{90\%} & \second{74\%} & 14\% & 14\% & \second{88\%} & \best{86\%} \\
\emph{Place Object Basket} & 43\% & 36\% & 44\% & 39\% & \second{74\%} & \best{75\%} & 37\% & 27\% & \best{89\%} & \second{73\%} \\
\midrule
\textbf{Average (\%)} & 42.98 & 43.84 & 72.80 & 72.84 & \second{77.56} & \second{77.00} & 41.32 & 39.90 & \best{80.44} & \best{81.48} \\
\bottomrule
\end{tabular}
}
\end{table*}

%% file: sections/05-real-world-experiments-arxiv.tex
\begin{figure*}[!htbp]
\centering
\includegraphics[width=0.99\textwidth]{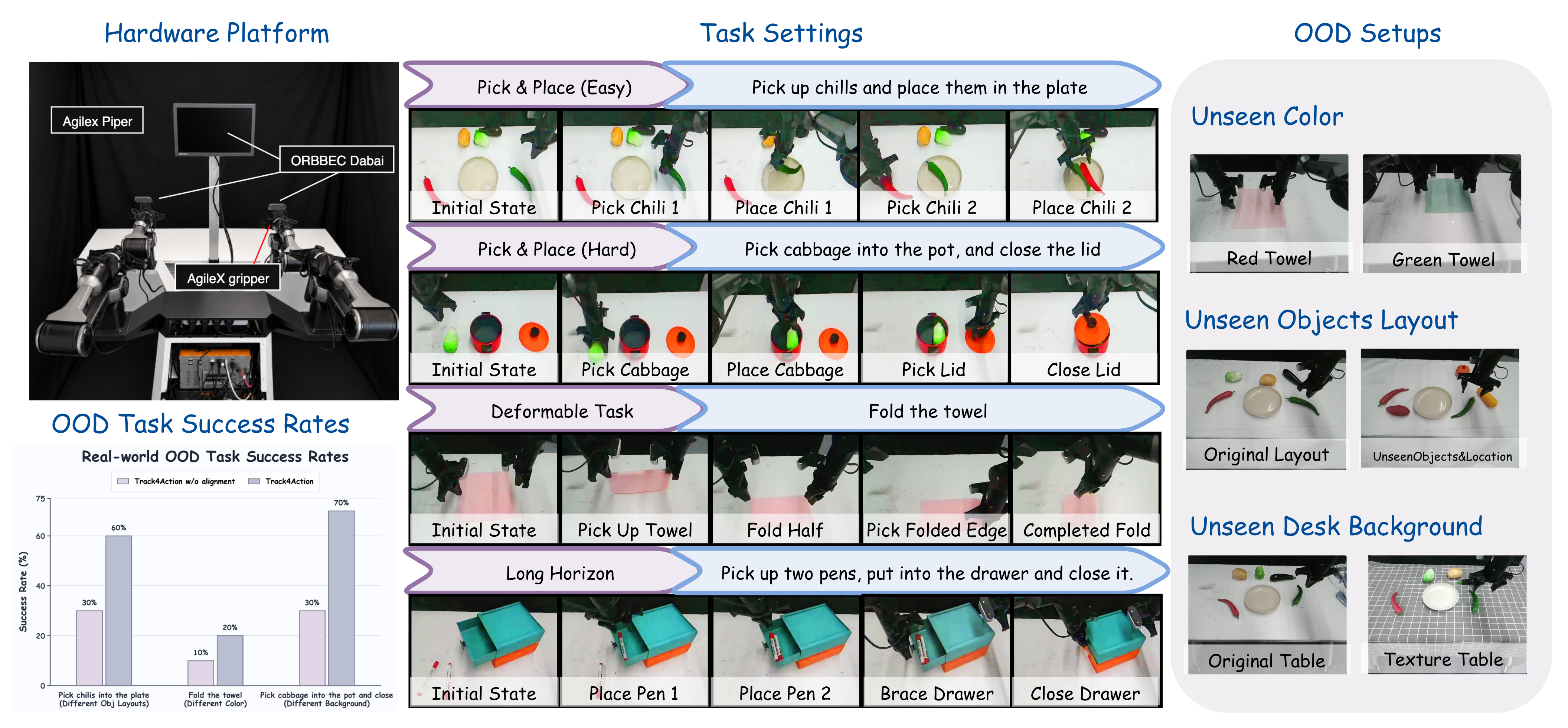}
\caption{\textbf{Illustration of Real-world evaluation.}
\textbf{Upper left:} the AgileX ALOHA-style hardware platform, built with two
Piper arms and equipped with three cameras: one front-facing camera and one wrist-mounted camera on each arm.
\textbf{Center:} four real-world manipulation tasks covering rigid-object
transfer, deformable-object manipulation, and long-horizon interaction.
\textbf{Right:} the OOD settings, including unseen towel color, object layout,
and desk background.
\textbf{Lower left:} OOD experimental results comparing alignment-free variant (w/o alignment)
with \method{} under the three settings.}
\label{fig:real_world}
\vspace{0.8\baselineskip}
\end{figure*}

\section{Real-World Experiments}
\label{sec:real_world}
The final question asks whether tracker supervision remains useful when the
policy leaves simulation and executes without \trackteacher{}.
We evaluate \method{} on physical bimanual manipulation under the same
training-only teacher and tracker-free deployment protocol.
Figure~\ref{fig:real_world} summarizes the platform, task families, and
out-of-distribution settings.

\paragraph{Setup.}
We conduct physical experiments on the AgileX ALOHA-style platform, which
comprises two 6-DoF Piper arms with parallel-jaw grippers.
One front-facing camera and two wrist-mounted cameras provide the RGB
observations.
The policy predicts a 14-D bimanual action composed of six joint commands and
one gripper command for each arm.
As in simulation, deployment runs only the VLA policy and does not execute
\trackteacher{}.

\paragraph{Tasks and metrics.}
We train \method{} on 50 demonstrations for each of four tasks and evaluate
every compared policy for 10 trials per task.
The suite contains two rigid-object tasks, \emph{place two chilies on a plate}
and \emph{place a cabbage in a pot and close the lid}; one deformable task,
\emph{fold a towel}; and one long-horizon task, \emph{place two pens in a
drawer and close it}.
We report \emph{success rate}, the fraction of trials that complete the full
instruction, and \emph{process score}, which awards 25 points for each of four
ordered task milestones and reaches 100 when all stages are completed. 
The milestone definitions are in Figure~\ref{fig:real_world}.
We compare with $\pi_{0.5}$ \citep{black2025pi05} and the alignment-free
variant.

\paragraph{Robustness protocol.}
We evaluate three held-out physical changes: a new object layout with
distractors for chili transfer, an unseen towel color, and a textured desk
background.

\paragraph{In-Domain Tasks and Process Score Details.}

We select four tasks: \emph{pick up the chilies and put them into the plate},
\emph{fold the towel}, \emph{put two pens into the drawer and close it}, and
\emph{pick up the cabbage, put it into the pot, and close the lid}.
These tasks cover rigid pick-and-place, deformable manipulation, coordinated
bimanual control, and multi-stage interaction.

\emph{Pick up the chilies and put them into the plate.} 
This task requires the robot to pick up the green and red chilies and place them into the plate, ignoring other vegetables. The task is separated into 2 steps: first, the robot needs to approach the fist chilies and pick them up, then the robot needs to find the plate and place the chilies right there. This task is relatively simple, it shows the robot's ability to understand the instruction and perform the correct action. For each chilies, pick and put respectively value 25 points, total score is 100 points.

\emph{Fold the towel.} 
This task requires the robot to pick up the towel and fold it into a neat rectangle. The task is separated into 3 steps: first, the robot needs to approach the right towel edge, grasp, and pick it up simultaneously with both of its 2 arms; then the robot needs to find a suitable surface to fold it on, and lay the towel down on it; finally, the robot needs to fold the towel properly while avoiding collisions. Simultaneous grasping values 25 points, lay down values 25 points, another folding values 25 points, and the last 25 lays on the towel's final state being flat and regular. This task tests the robot's ability to handle deformable objects and perform precise manipulation tasks. 

\emph{Pick up the two pens, put them into the drawer, and close it.} 
This task requires the robot to pick up the two pens on table and put them into the drawer one by one, then close the drawer with 2 arms. The task is separated into 3 steps: first, the robot needs to figure out the positions of the pens, decides the order, and pick them up; then the robot needs to find the drawer's open layer and put the pens into it; finally, the right arm should hold the suitable place on the drawer, and then the left arm close the drawer properly. Both of the 2 pens' pick and place task value 25 points, the right arm hold the drawer values 25 points, and the left arm last closing the drawer values 25 points. This task tests the robot's ability to perform coordinated bimanual manipulation.

\emph{pick cabbage into the pot and close the lid.} 
This is also a Pick\&Place task, requires the robot to pick up the cabbage, put it into the pot, then close the lid properly. The task is separated into 3 steps: first, the robot needs to find the cabbage and pick it up; then the robot needs to locate the pot and put the cabbage into it; finally, the robot should grasp the lid's handle and close the pot properly. First pick and put tasks respectively values 25 points, second pick task values 25 points, and robot closing the lid without collision or dropping values 25 points.

\paragraph{Out-of-Domain Tasks}

We set 3 conditions for OOD test.

\emph{Unseen Colors.} We change the color of the objects in the scene, for example we use a green towel instead of a red one as in the training dataset, This tests the robot's ability to generalize to unseen object colors.

\emph{Unseen Objects Layout.} We change the layout of the objects in the scene, as well as the location and rotation of the target operating object, and let the robot predict the appropriate actions for each scenario, testing the robot's ability to generalize to unseen object layouts.

\emph{Unseen Background.} We change the background of the operating scene, by covering the desk with tablecloth of different color and texture. Testing the robot's ability to generalize to unseen background conditions.

\begin{figure}[!t]
\centering
\includegraphics[width=\columnwidth]{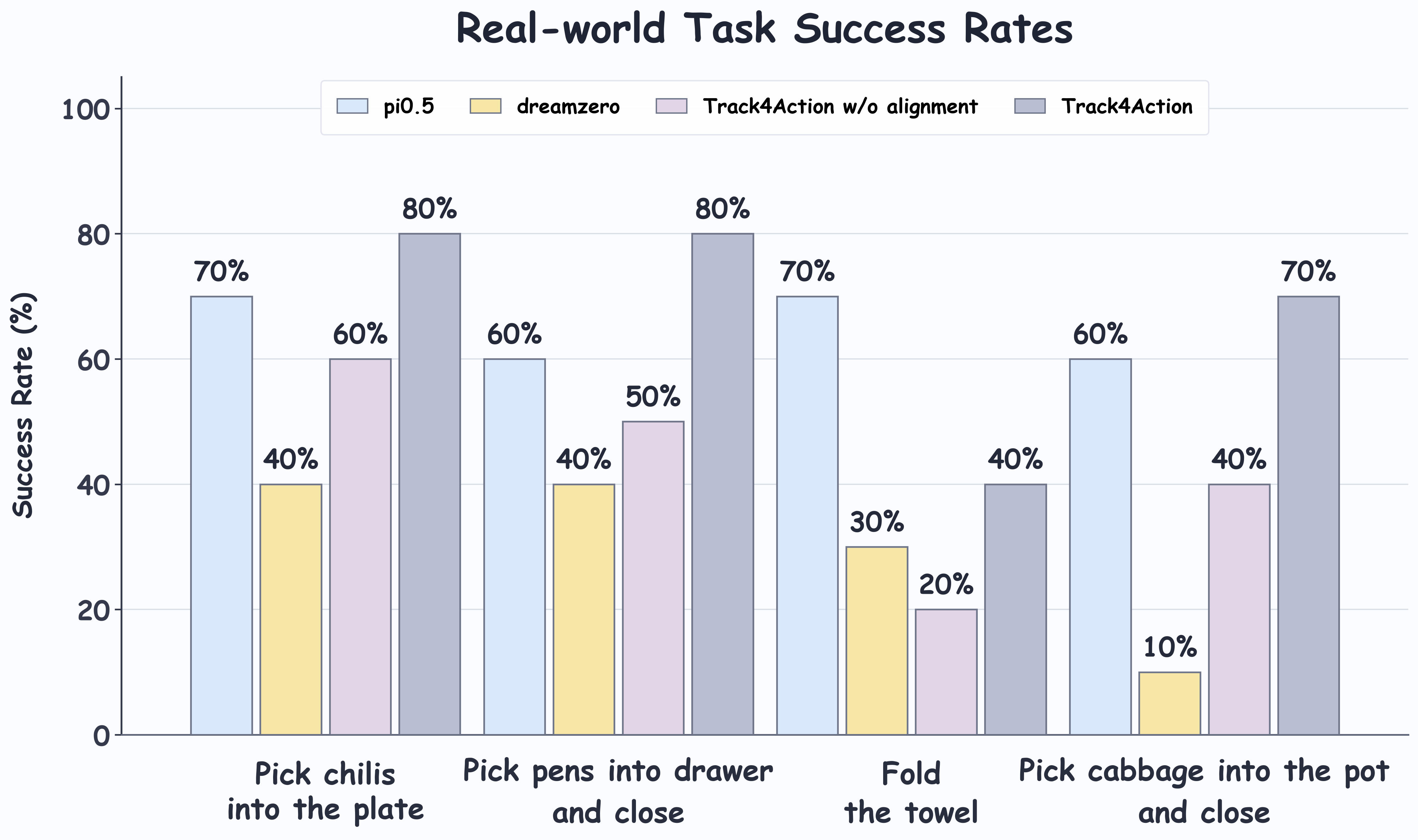}
\caption{\textbf{Real-world task success rates.}
Success rate measures full task completion over 10 trials per task.
Across the four tasks, \method{} averages 67.5\% success.
The ``w/o alignment'' legend denotes the variant trained without tracker
supervision.}
\label{fig:realworld_success}
\end{figure}

\begin{figure}[!t]
\centering
\includegraphics[width=\columnwidth]{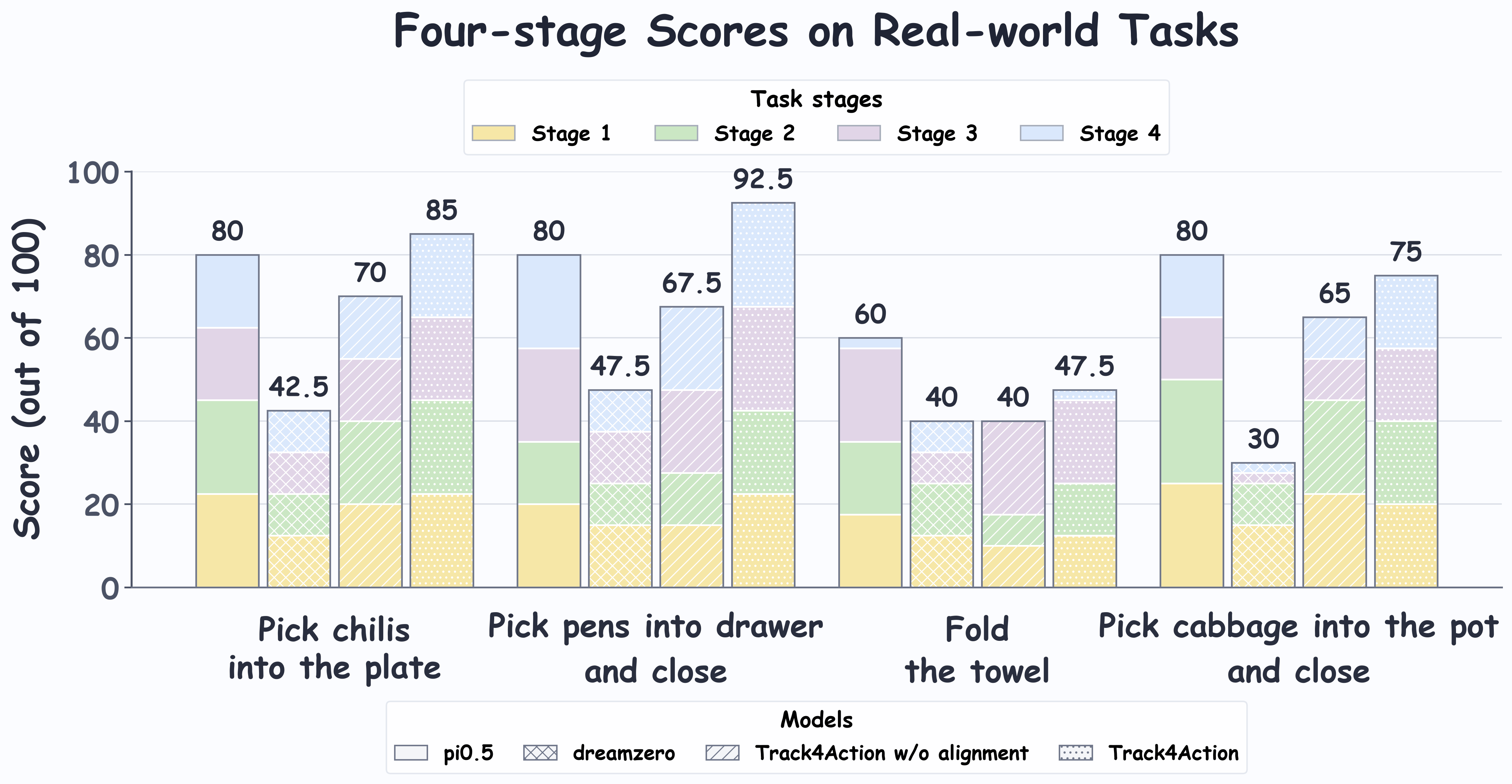}
\caption{\textbf{Four-stage process scores on the real-world tasks.}
Each stacked bar decomposes the process score into four ordered, equally
weighted milestones; its total height is the resulting score out of 100.
Across the four tasks, \method{} averages 75.0.}
\label{fig:realworld_stages}
\end{figure}

\paragraph{In-distribution results.}
Figure~\ref{fig:realworld_success} shows that \method{} reaches 67.5\% average
success, compared with 65.0\% for $\pi_{0.5}$ and 42.5\% for the
alignment-free variant.
Tracker supervision improves success on every task by at least 20 points over
the alignment-free control.
The largest 30-point gains occur on the drawer and cabbage tasks, both of which
require completing several ordered stages.

Figure~\ref{fig:realworld_stages} shows that the average process score of
\method{} is 75.0, equal to $\pi_{0.5}$ and 14.4
points above the 60.6 score of the alignment-free variant.
Relative to the alignment-free control, the process score increases on all four
tasks, with gains ranging from 7.5 to 25.0 points.
The comparison with $\pi_{0.5}$ is less uniform.
\method{} improves drawer success from 60\% to 80\%, but reaches only 40\% on
towel folding compared with 70\% for $\pi_{0.5}$.
This failure identifies deformable manipulation as a remaining physical
challenge despite the higher average completion rate.

\paragraph{Out-of-distribution results.}
The left panel of Figure~\ref{fig:real_world} reports tracker-free execution
under the three held-out changes.
Tracker supervision raises chili-transfer success from 30\% to 60\% under the
new object layout, towel-folding success from 10\% to 20\% for the unseen
color, and cabbage-task success from 30\% to 70\% on the textured desk.
The average across these settings increases from 23.3\% to 50.0\%.
The 20\% result on the unseen-color towel also shows that physical
out-of-distribution deformable manipulation remains difficult.

%% file: sections/06-limitations.tex
\section{Limitations}
\method{} requires temporally ordered demonstrations because its training
target reads the realized clip $V_{t:t+K}$.
Isolated image-action pairs cannot provide this supervision, and independently
moving objects can weaken the association between a robot command and the
observed scene transition.
The pooled target emphasizes segment-level world motion and may discard local
contact correspondence.
It also inherits errors from \trackteacher{} under occlusion or fast motion and
uses only the primary camera view.
These choices add offline preprocessing but no deployment cost.
Our physical evaluation covers one bimanual platform, four tasks, and three
visual shifts.
Evaluating broader embodiments, camera configurations, and physical
disturbances remains an important next step.

%% file: sections/07-conclusion.tex
\section{Conclusion}
We asked whether the realized 3D transition in a demonstration can supervise a
VLA without becoming a deployment input.
\method{} answers this question by distilling an action-aligned
\trackteacher{} feature into track queries that also condition the action head
through gated fusion.
The teacher feature only defines the training target, and the deployed policy
remains tracker-free.
\method{} reaches 82.3\% on zero-shot LIBERO-Plus, 80.44\% and 81.48\% on the
clean and randomized RoboTwin 2.0 splits, and 67.5\% average success across four
physical bimanual tasks.
The cross-domain results show that demonstration video can supervise both the
command a policy executes and an action-facing representation of the world
transition associated with that command.